# Security Fix Dynamics: the Case of the Linux Kernel

Nicholas Beale, Sciteb with substantial AI assistance[1]

**Abstract.** If AI destabilises cybersecurity, it is important to understand the dynamics of the processes by which vulnerabilities are detected and fixed. Using history data of the Linux kernel, this paper measures the size of fixes, the rate at which code of a given age is found to need fixing, and how that rate changed in 2026 as AI tools came into wide use. From 2018 to 2025 the process was stable. Fixes were small, and no larger for bugs that had taken years to find; the expected number of follow-up fixes rose with the size $s$ of a commit as $s^{\gamma}$ with $\gamma \approx 0.7$; follow-ups accumulated in proportion to the logarithm of elapsed time; and the rate at which code of each age attracted fixes was nearly constant. From about the start of 2026 that rate rose for all code more than six months old, in earlier fixes and ordinary code alike, by a factor of about three within nine months after allowing for more complete recording. The rise was greatest for older code, and extended to the kernel's most scrutinised subsystems. A branching-process and queueing model shows that if this multiplier were sustained the five-year re-fix ratio would nearly double, to about 0.17, and remediation demand would be about two and a half times its historical level. The model also gives the conditions for the re-fix process to become self-perpetuating, and shows that demand becomes severe well before that point is reached.

## 1 Introduction

A companion paper [Bea26] argues that AI destabilises the contest between attackers and defenders of software, so that patching faster cannot by itself restore security. That argument is based on theoretical considerations including especially economic game theory. But it raises the adjacent question of the underlying statistical properties of vulnerabilities are fixed in practice.

The Linux kernel is an attractive public dataset on which to ask the question. Every change is recorded; a fix conventionally names the commit it repairs in a `Fixes:` tag; and since 2024 the kernel has assigned a CVE to every fix with a plausible security consequence. In some respects it is a best-case scenario, and should not be assumed to be typical of all software. It is a high-maturity, high-scrutiny benchmark: the most reviewed and most fuzzed[2] codebase in open source, with a design inherited from Unix, so arguably a best case on process and design maturity. On language it is not: C, concurrency and hardware interaction make some classes of defect likelier than in memory-safe code. It would be very interesting to run similar analyses on other code bases to compare and contrast.

The paper makes six contributions.

1. A description of fix size on the full population, which shows no evidence of variation with the age of the bug.

---

[1] The text of the paper was almost entirely written by Claude Fable 5.2 under my direction, and Fable 5.1 and 5.2 did all the data acquisition and analysis. Drafts 3, 5 and 7 were peer reviewed by GPT-6.2, which provided suggestions for additional empirical analyses, statistical robustness tests and extensions to the mathematical model. Gemini Neon peer reviewed drafts 6 and 7 and also provided valuable suggestions.

[2] Fuzzing is automated testing that feeds a program millions of random or mutated inputs and watches for crashes. Coverage-guided fuzzing (syzkaller, for the kernel, since 2015) keeps the inputs that reach new code and mutates those, so it finds bugs in proportion to how much of the code it can reach.

2. An empirical re-fix law: expected follow-ups grow as a power 0.7 of fix size, broadly stable across a decade and across security fixes, other fixes and ordinary commits.
3. The dynamics of re-fixing: follow-ups share the size distribution of fixes and accumulate logarithmically in time, at age-specific rates that were constant from 2019 to 2025.
4. The 2026 break: code more than six months old is being fixed at a rising multiple of its historical rate, in earlier fixes and ordinary commits alike, with evidence from the kernel's own records on who is producing the fixes and with what tools.
5. A model using these regularities that estimates: the mean and variance of total remediation work, waiting time under load, and the effect of faster discovery on arrivals and on re-fixing.
6. The observation that the next 6-9 months should allow clearer trend identification and distinguish between statistical regimes with fundamentally different implications.

## 2 Related work

**Security patch size.** Li and Paxson [LP17] examined more than 4,000 security fixes across 682 open-source projects, finding them smaller and more local than other bug fixes, with a median of 7 lines changed. They estimated that approximately 7% of security fixes were incomplete and 5% regressive, and investigated the intervals before subsequent correction. Their median security fix size matches that found here.

**Commit size.** Arafat and Riehle [AR09] and Kolassa et al. [KRS13] investigate the heavy-tailed distribution of commit sizes in open-source software. Gorshenev and Pis'mak [GP04] report power-law probability distributions with exponents near −1.5 for numbers of lines added or deleted in individual file modifications. Clauset, Shalizi and Newman [CSN09] establish statistical methods for testing power-law claims; Appendix A follows their emphasis on likelihood-based estimation and comparison with alternative distributions.

**Bug-introducing changes and incorrect fixes.** Śliwerski, Zimmermann and Zeller [SZZ05] introduced methods for tracing fixes to the changes that induced them and investigated the association between change size and subsequent defects. The kernel's Fixes: tag provides an explicitly recorded link. Yin et al. [YYZ11] found that approximately 15–24% of sampled fixes for post-release bugs in large operating systems were incorrect; Jones [Jon12] estimates the industry rate of new defects introduced by repairs at about 7%. Park et al. [PKRB12] investigate supplementary fixes in three open-source projects. Liu et al. [LZJ25] characterise 26 instances of incomplete security fixes in the Linux kernel, including their size, timing and root causes; Qi, Li and Wang [QLW26] analyse 1,646 multi-patch vulnerability remediation records. Yomtov [Yom26] reports a substantial rise in recorded broken security fixes in 2026 and finds that 3.2% of kernel CVEs were introduced by commits fixing other CVEs. The present study estimates the joint dependence of expected follow-up counts on commit size, age and calendar period across a large population of security and non-security commits.

**Queueing and self-excitation.** Branching-process models of imperfect computer repair date at least to Lewis [Lew64], and subsequent software reliability models have incorporated the generation of new defects during debugging. Hawkes [Haw71] and Hawkes and Oakes [HO74] provide the mathematical framework for self-exciting processes and their representation as age-dependent immigration–birth processes. More recently, Yun et al. [YES26] model the dynamics of vulnerability discovery, exploitation and patching, finding heavy-tailed patching times; Faghani [Fag26] models the effects of AI-accelerated discovery on finite-capacity remediation and systemic compromise; and Stucke [Stu26] examines remediation as a capacity-allocation problem. The present model combines empirically estimated fix-size and follow-up distributions with branching and queueing theory to investigate the effects of accelerated discovery on total remediation demand.

## 3 Data and methods

**Sources.** The mainline kernel repository at commit *25faec9ce61c* (1,508,451 commits) and the CVE List V5 at commit *492d626d3a68* (26 November 2026), from which the 19,202 records assigned by the kernel's CVE Numbering Authority were taken.

**Populations.** Three populations were used, all measured the same way.

1. *CVE fixes*: 18,726 distinct mainline commits that fix a published kernel CVE, authored 2021–2026 (one changes no source lines and is left out of size statistics). The kernel became a CVE authority in February 2024 and assigned CVEs retrospectively to earlier fixes, so pre-2024 cohorts were selected with hindsight.
2. *Tagged fixes*: all 146,703 mainline commits carrying a `Fixes:` tag that resolves to a mainline commit; 141,406 are measurable and dated 2014–2026. CVE fixes are a subset.
3. *Ordinary commits*: non-merge commits without a `Fixes:` tag. Analyses that need sizes use a random sample of 60,000 from 2021–2025; analyses of rates by age and period (Section 6) use all of them, 733,000 since October 2015, with commits back to 2006 used to establish normal rates at high ages.

**Size** was estimated as source lines added plus deleted, excluding documentation, self-tests and text files. Commits above 100 lines in which deletions exceed 90% of the change are removals of whole drivers or filesystems (21 CVE fixes; the largest deletes 168,000 lines) and are excluded; Appendix B gives the sensitivity to this exclusion. Regressions also exclude the few commits above 1,000 lines which would otherwise be distortive.

**Follow-ups.** A follow-up to a commit is a later mainline commit whose `Fixes:` tag names it. Counts are taken within a horizon $H$ of the commit's author date, and at each horizon only commits at least $H$ old on the snapshot date are included, so cohort sizes fall as $H$ rises. The measure is a lower bound, since a follow-up without the tag is missed, and its coverage rose during the 2010s as the tag was adopted.

**Dates and multiple targets.** Ages are measured between author dates; Appendix C repeats the main analysis on commit dates. A fix may name more than one target (3.9% do) and is then counted as a follow-up to each, so follow-up counts slightly exceed the number of distinct repairing commits.

**Bug age.** Author date of the fix less author date of the commit that the CVE record names as introducing the bug.

**Patch series.** A fix may be one of several commits submitted together. Series membership is not recorded in the repository; it is approximated by runs of commits with the same author and committer applied within five minutes of each other. On this proxy 61% of CVE fixes stand alone and 39% belong to a series.

**Throughput, not arrivals.** A CVE is assigned after the fix exists. All counts here are of completed fixes: they measure the output of the remediation process, not the arrival of reports. The two differ exactly when the process is under strain (Section 6).

The distribution of the fixes designated as CVE by quarter is shown below in Fig 1, which clearly indicates that something changes in 2026.

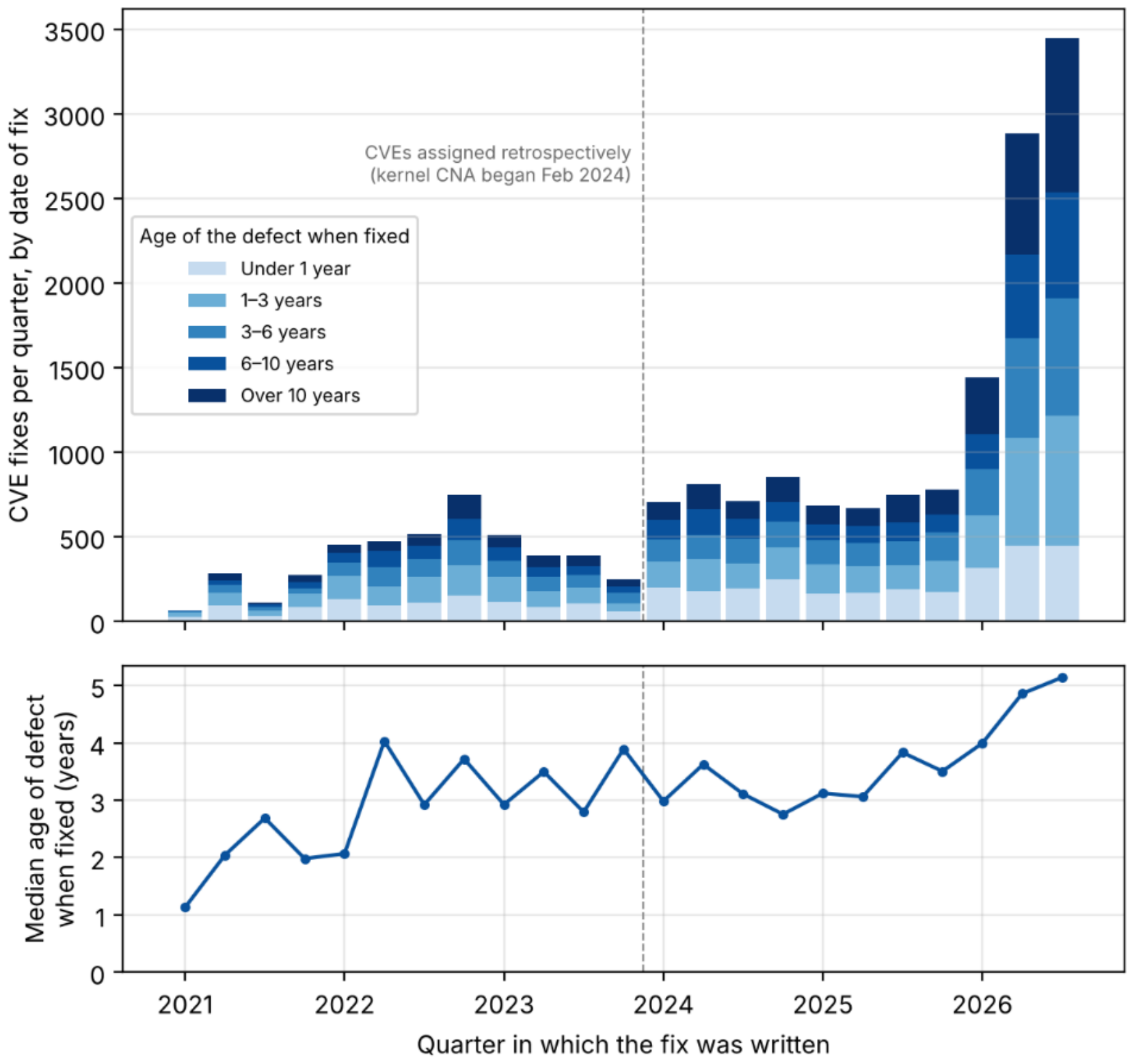

*Figure 1. Number of CVE Fixes, also showing age of defect when fixed. May undercount in Q3 2026.*

It is worth noting that the assignment of a CVE the date of the bug that CVE fixes is sometimes made after the fix is deployed, so the final bar in Q3 2026 may well be higher eventually. It will be especially interesting to see whether there is any sign that the trend is such as to allow even a rough estimate of the total number of CVEs in the Linux Kernel. At present it is not: about 3,000 a quarter are now being fixed, against 750 before 2026 and there are certainly plenty more to find.

## 4 The size of a fix

The median CVE fix changes 7 source lines; the mean is 15.7; 1.8% exceed 100 lines; the squared coefficient of variation is 3.9. Tagged fixes in general are the same (median 6, mean 14–15 in every year from 2014 to 2025). The distribution of fix sizes is shown in Figure 2

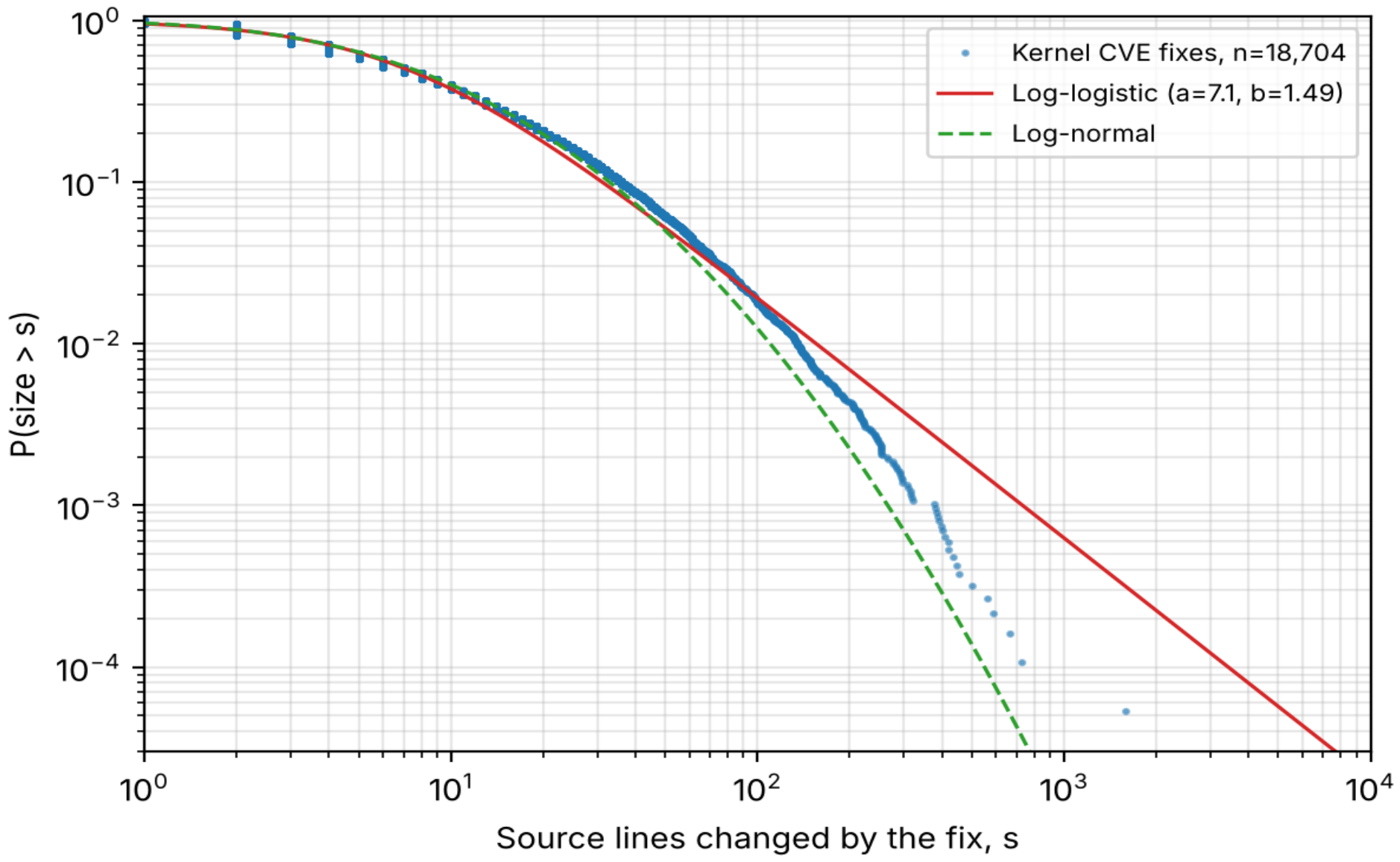

*Figure 2. Complementary distribution of source lines changed by kernel CVE fixes, with fitted log-logistic and log-normal laws.*

Three observations arise:

1. **No simple law fits body and tail.** The body is close to log-normal. The tail is heavier than log-normal up to about 300 lines and lighter beyond. A truncated-Pareto fit gives a local exponent that rises steadily with the threshold, from 1.3 above 20 lines to 1.8 above 50, so there is no stable power law (Appendix A). The model in Section7 below uses empirical moments only.
2. **Fix size is flat in bug age.** The median fix is 7 lines for bugs under a year old, 7 for one to four years, and 7 for older (90th percentiles 35, 36, 36). The commits that introduced the bugs are far larger: median 156 lines.
3. **A commit is not the whole remedy.** Summing over the series a fix belongs to, the median rises from 7 to 12 lines, the mean from 16 to 51 and the squared coefficient of variation from 3.9 to 14; 2.8% of series exceed 300 lines. The kernel's convention of small, separable patches bounds the size of a commit, not the work of a remediation. The series proxy is crude and may include unrelated patches by the same author, so these are upper estimates.

# 5 When a fix does not hold

## 5.1 Approximate re-fix law

Figure 3 and Table 1 give the mean number of follow-ups within two years by size of fix, for the 101,161 tagged fixes of 2014–2024.

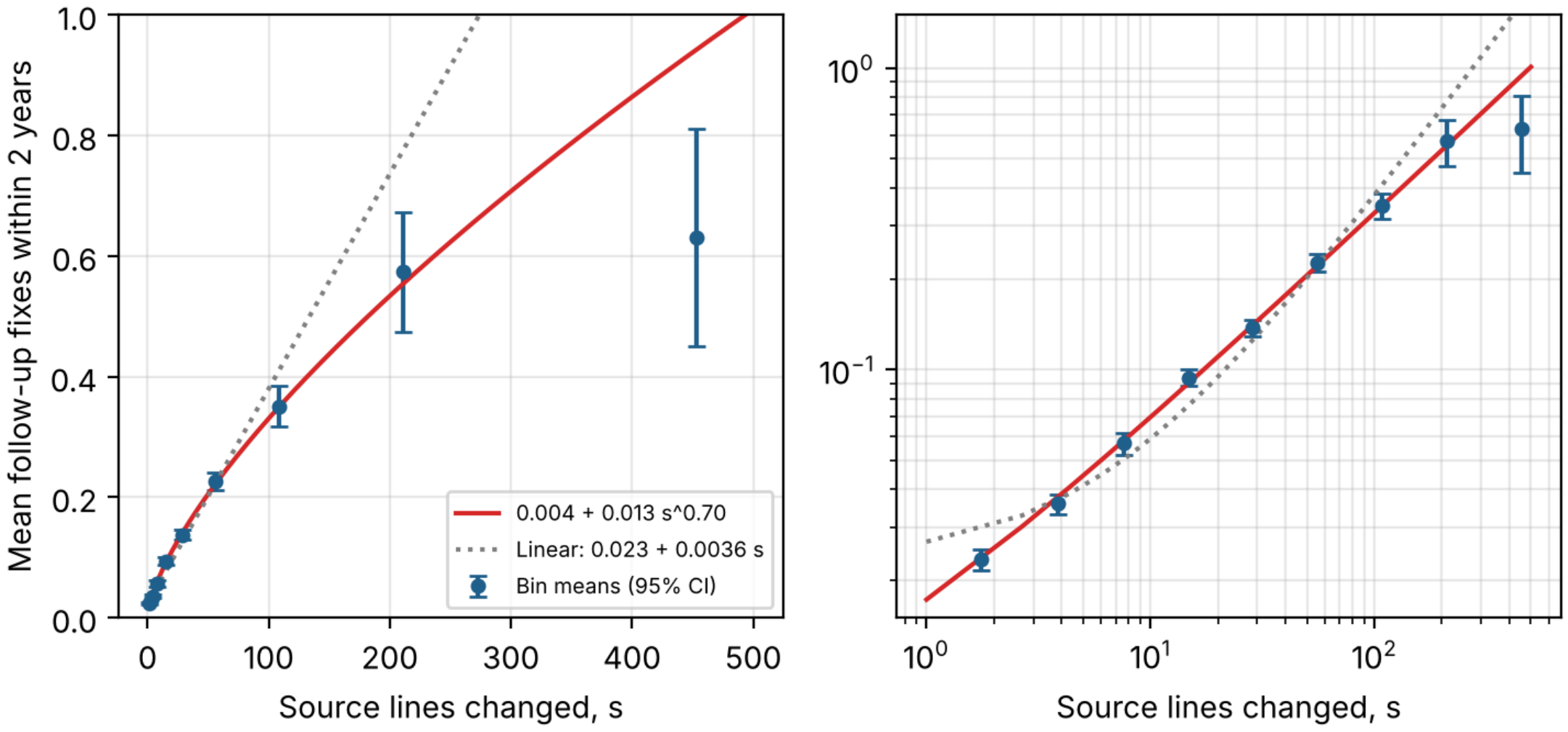

*Figure 3. Mean follow-up fixes within two years against fix size (bin means with 95% intervals), tagged fixes 2014–2024, on linear and logarithmic axes, with the fitted power law and the best linear fit.*

*Table 1. Follow-up fixes within two years by fix size, tagged fixes 2014–2024.*

| Lines changed | n | Mean lines | Mean follow-ups |
|---|---|---|---|
| 1–2 | 28,535 | 1.7 | 0.023 |
| 3–5 | 21,745 | 3.9 | 0.036 |
| 6–10 | 18,225 | 7.7 | 0.057 |
| 11–20 | 14,963 | 14.8 | 0.094 |
| 21–40 | 9,962 | 28.5 | 0.138 |
| 41–80 | 5,177 | 56 | 0.226 |
| 81–160 | 1,939 | 109 | 0.351 |
| 161–320 | 493 | 211 | 0.574 |
| 321–1,000 | 122 | 453 | 0.631 |

Write $N_H(s)$ for the expected number of follow-ups within horizon $H$ to a fix of $s$ lines. Fitted by Poisson maximum likelihood, we have the following approximate equation:

$$N_H(s) \approx c_0 + c_1\, s^{\gamma}, \qquad [1]$$

$$\gamma = 0.70\ (0.66\text{–}0.76), \quad c_0 = 0.004, \quad c_1 = 0.013 \quad (for\ H = 2\text{ years}).$$

A linear law $m + d\, s$ ($m = 0.023$, $d = 0.0036$ per line) is a fair approximation below 200 lines but is rejected against the power law ($\Delta\mathrm{AIC} = 155$) and over-predicts large fixes. The exponent is 0.66–0.71 at every horizon from 30 days to three years, 0.72–0.76 for ordinary commits, and 0.71 (s.e. 0.01) in a joint model of size, age, period and population (Appendix C). For CVE fixes alone it is 0.8–0.9 with a wide interval (0.70–0.93). Sublinearity is robust across all three populations; a common exponent is not established. Even the power law over-predicts the largest bin (0.95 against 0.63 observed at 453 lines, on 122 fixes), so extrapolation beyond a few hundred lines is unsafe. The intercept $c_0$ is small: a two-line fix is followed up 2.3% of the time, and most of that is the size term.

Three qualifications are necessary.

1. **The decomposition is statistical.** Classifying follow-ups as incomplete or regressive, as Li and Paxson did, is needed before a mechanism is claimed.

2. **Counts are over-dispersed.** A negative-binomial model with variance $\mu + \alpha\mu^2$ is preferred to Poisson ($\alpha = 3.3$ on CVE fixes), and reproduces the share of fixes with at least one follow-up at every size, which the Poisson model overstates for large fixes (30% predicted against 21% observed above 80 lines). Means are unaffected; confidence intervals clustered by subsystem (363 clusters) are within 5% of the unclustered ones.

3. **Larger fixes attract fewer recorded follow-ups per line.** The rate per line falls as $s^{-0.3}$. Whether they introduce fewer defects per line is a further step: one follow-up may repair several defects, and large changes may be reviewed harder or contain more repetitive code. The number of functions, files or hunks changed may prove a better measure than lines, and is a natural next test.

## 5.2 Fixes against ordinary code

Table 2 compares the three populations over identical cohorts and windows.

*Table 2. Follow-ups per commit, cohorts of 2021–2025.*

| Population | n (180 d) | Median size | Mean size | Follow-ups per commit, 180 d | 2 years |
|---|---|---|---|---|---|
| Ordinary commits | 53,156 | 15 | 48 | 0.076 | 0.124 |
| Tagged fixes | 66,982 | 6 | 14.5 | 0.052 | 0.077 |
| CVE fixes | 7,929 | 6 | 13.8 | 0.060 | 0.089 |

Ordinary commits are followed up more often because they are larger. At equal size the order reverses: in every size class up to 320 lines a fix attracts 1.1 to 1.5 times as many follow-ups as an ordinary commit of the same size. Fix code is somewhat more likely to need fixing than ordinary code, line for line; the joint model of Appendix C puts the ratio at 1.14 at equal size and age.

## 5.3 Follow-ups seem to have the same size distribution

The 1,528 follow-ups to CVE fixes have median 7 lines, against 7 for CVE fixes in general, and the two distributions are not distinguishable by a Kolmogorov–Smirnov test ($p = 0.46$), which is absence of evidence of a difference rather than proof of none. Size is weakly inherited: the elasticity of follow-up size on parent size is 0.16 (s.e. 0.02), so a hundredfold larger parent has a follow-up about twice as large. Classing fixes as small (up to 10 lines), medium (11–50) and large, the matrix of mean offspring by type (Appendix C) has spectral radius 0.066 at 180 days against a single-type ratio of 0.061, so the dependence adds under a tenth to the re-fix ratio.

## 5.4 Follow-ups accumulate with the logarithm of time

Figure 4 plots the follow-ups per fix, with the data divided into 12-month periods from 1 Oct to 30 Sept, to avoid distortion caused by partial data in the most recent two months of data. It is evident that these curves used to be close to straight lines in log time, with something different about the first two years, but that something seems to have happened recently.

The two earliest cohorts lie lower because the $Fixes$ tag was less used then: the share of commits carrying it rose from 4% in 2015 to 12% in 2018 and was 15–18% from 2019 to 2025. From the 2017–18 cohort onward the curves coincide, at about 0.016 follow-ups per e-fold of time, until each turns upward at its right-hand end. Those upturns all fall in calendar 2026 and are analysed in Section 6. Rates here are per commit, not per surviving line: no allowance is made for the parts of an old commit that have since been rewritten or deleted, and such attrition is one possible contributor to the $1/t$ decline. On data before 2026 the log-time law holds to about five years; beyond that the old cohorts' tails are inflated by the 2026 effect and the law's range cannot be judged.

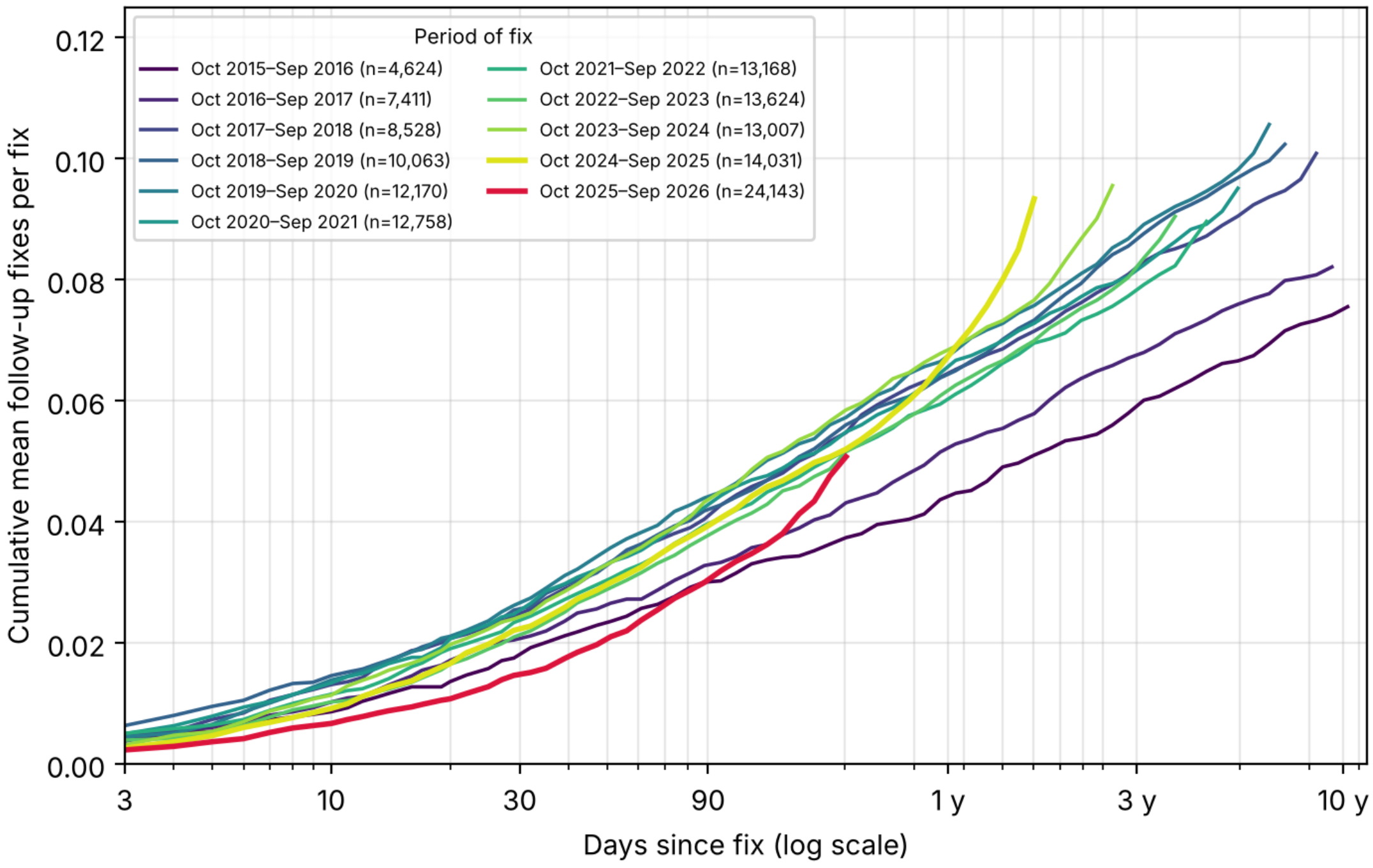

*Figure 4. Cumulative mean follow-ups per fix against days since the fix (log scale), by year of fix (October to September), observed to 30 September 2026.*

The consequence is that the re-fix ratio has no single value. Table 3 gives it by horizon.

*Table 3. Re-fix ratio by horizon, tagged fixes of 2018–2023 (n = 71,206; 69,987 at three years, since fixes of December 2023 are not yet three years old). Parameters are from equation [1]*

| Horizon $H$ | 30 d | 90 d | 180 d | 1 y | 2 y | 3 y |
|---|---|---|---|---|---|---|
| $R_H$ (follow-ups per fix) | 0.024 | 0.041 | 0.052 | 0.064 | 0.076 | 0.084 |
| $c_1$ | 0.0044 | 0.0069 | 0.0088 | 0.0105 | 0.0126 | 0.0142 |
| $\gamma$ | 0.66 | 0.69 | 0.70 | 0.71 | 0.71 | 0.70 |

The exponent is basically constant and the scale $c_1$ carries the whole dependence on time. Calendar days are also the wrong clock if the pace of work changes; mainline commit volume was 84,000–93,000 a year throughout 2023–2026, so for the kernel the two clocks still agree.

## 5.5 Broadly stable from 2018 to 2023

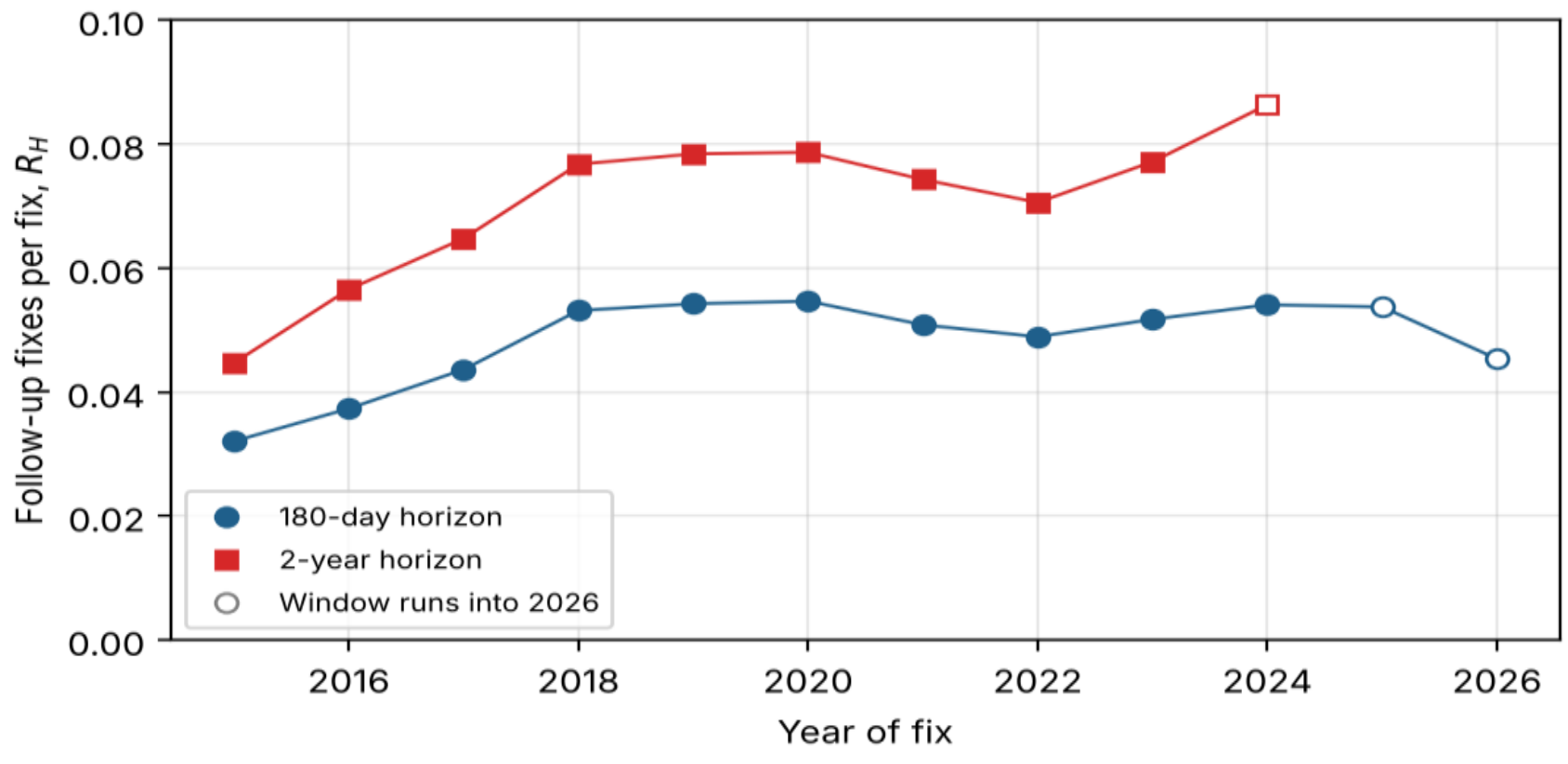

*Figure 5. Re-fix ratio of tagged fixes by year of fix, at 180-day and two-year horizons. Hollow markers are cohorts whose observation window runs into 2026.*

Both series rise by about 70% from 2015 to 2018. That is probably due to the spread of the `Fixes` tag rather than more re-fixing, but the data cannot separate the two. From 2018 to 2023 both are flat: 0.049–0.055 at 180 days and 0.071–0.079 at two years, with the two-year figure a constant 1.45–1.48 times the 180-day one, consistent with an unchanging log-time law. The fix quality, so measured, did not improve over those six years. The hollow points depart in opposite directions, the two-year series upward and the 180-day series downward; both are effects of 2026 and are taken up in Section 6.

# 6 The break of 2026

## 6.1 Throughput and age

CVE fixes ran at about 3,000 a year in 2024–25 and reached 8,283 in the first eleven months of 2026. Tagged fixes of all kinds ran at about 1,200 a month through 2025 and at 2,300–3,200 a month from April to September 2026. Fixes marked for backporting to stable kernels, which are identified without reference to the `Fixes:` tag, rose 3.4-fold over the same period. These are completed fixes; reports may have risen by more. The median age of the bug fixed by a CVE fix rose from 3.1–3.4 years in 2022–25 to 4.9 years in 2026, and the share older than ten years from about 16% to 25%. Fix size did not change (monthly median 7–9 lines).

## 6.2 Age-specific rates were constant, then moved

Table 4 gives the rate at which tagged fixes attract follow-ups, by the age of the fix when the follow-up was written and by the calendar period in which it was written.

*Table 4. Follow-ups per 1,000 fix-years at risk, by age of fix and calendar period.*

| Age of fix | 2019 | 2020 | 2021 | 2022 | 2023 | 2024 | 2025 | 2026 Jan–Apr | 2026 May–Sep |
|---|---|---|---|---|---|---|---|---|---|
| 0–30 days | 314 | 327 | 286 | 267 | 268 | 278 | 275 | 247 | 133 |
| 30–180 days | 69 | 71 | 67 | 65 | 67 | 80 | 69 | 82 | 72 |
| 180 days–1 year | 26 | 21 | 25 | 23 | 22 | 29 | 23 | 33 | 52 |
| 1–3 years | 9.5 | 9.6 | 10.9 | 9.2 | 9.2 | 8.8 | 9.7 | 17 | 31 |
| 3–5 years | 3.9 | 4.5 | 3.5 | 4.1 | 4.3 | 4.5 | 4.8 | 6.5 | 11 |
| 5–11 years | — | 2.2 | 1.7 | 2.5 | 2.2 | 2.4 | 2.0 | 4.1 | 6.9 |

Every row is flat for seven years. In 2026 the rate for fixes more than six months old rose two- to fourfold, the rate for fixes under a month old halved, and the 30–180-day band did not move. October 2026 is excluded here and below because some follow-ups written then had not been merged at the snapshot.

## 6.3 The multiplier

Define the **Fix Rate Multiplier** $M$ for a body of code in a period as the number of fixes it received divided by the number expected from its age, at the age-specific rates of October 2018 to September 2025. Figure 6 shows it month by month for code written before October 2025.

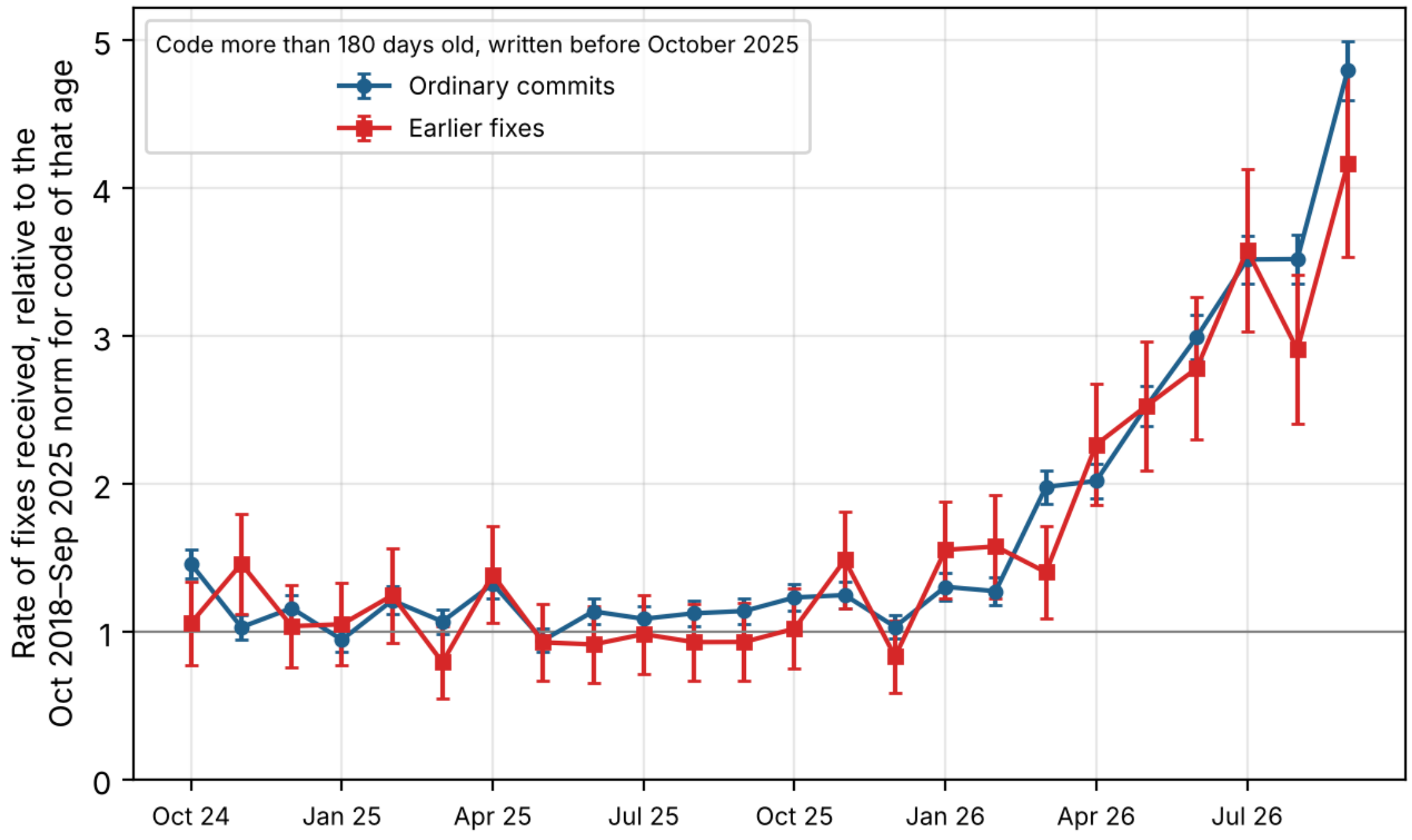

*Figure 6. Monthly rate at which code more than 180 days old received fixes, relative to the norm for its age, for ordinary commits and for earlier fixes, with 95% intervals.*

*Table 5. Fix Rate Multiplier by quarter, code written October 2015–September 2025 at ages over 180 days (95% intervals from a bootstrap over weeks of the quarter and years of the baseline).*

| Quarter | All code | Ordinary commits | Earlier fixes |
|---|---|---|---|
| Oct–Dec 2025 | 1.16 (0.93–1.38) | 1.17 (0.95–1.40) | 1.11 (0.83–1.44) |
| Jan–Mar 2026 | 1.52 (1.31–1.74) | 1.53 (1.29–1.78) | 1.51 (1.28–1.73) |
| Apr–Jun 2026 | 2.50 (2.27–2.77) | 2.50 (2.22–2.79) | 2.51 (2.25–2.74) |
| Jul–Sep 2026 | 3.89 (3.32–4.54) | 3.92 (3.29–4.59) | 3.53 (2.92–4.12) |

The third-quarter figures rest on 5,743 fixes to ordinary commits and 462 follow-ups to earlier fixes; the all-code column sums observed and expected counts over the two. The multiplier was within 0.8–1.15 in every year from 2018 to 2025. It rose from the turn of 2026 and reached 3.9 in the third quarter, with similar values for earlier fixes and for ordinary commits (the difference, 0.4, has a 95% interval of −0.5 to 1.3). The effect is therefore not about fixes: kernel code in general is being found defective at several times the historical rate, and fixes are part of that code.

How large the rise is depends on what it is compared with, because the recording of fixes was itself improving (Appendix C).

1. *Like-for-like baseline*. Against the most recent year alone, October 2024–September 2025, the third-quarter multiplier is 3.4 (2.9–3.8).
2. *Tagging*. Between that year and the third quarter of 2026 the share of independently identified fixes that carry the tag rose from 81% to 93%, a factor of 1.12–1.15. Allowing for it, the multiplier is about 3.0 (2.5–3.4). This is the figure used in the scenarios of Section 7.
3. *Dates*. Using commit dates rather than author dates changes the result by under 3%.

**Size.** There is weaker evidence that the rise was also greater for larger commits. In the joint model of Appendix C the size exponent, 0.71 before 2026, is higher by 0.16 in 2026 (s.e. about 0.02 allowing for over-dispersion). A direct test with controls gives about half that. Among fixes received at ages over 180 days by commits of known size, the exponent in the third quarter of 2026 exceeds that of 2024–25 by 0.09 (s.e. 0.04) after allowing for population, year of writing, subsystem and the experience of the original author; the controls do not reduce it. The rise is confined to the core subsystems (0.24, s.e. 0.08) and is absent in drivers (0.01, s.e. 0.06), and it does not differ significantly between fixes that declare AI assistance and those that do not. The size exponent in 2026 is therefore between about 0.80 and 0.87, and the evidence that the new methods are relatively better at finding defects in large changes is suggestive for core code and absent elsewhere.

**Cohorts.** Table 6 and Figures 7 and 8 show the same effect cohort by cohort for the twelve months to September 2026.

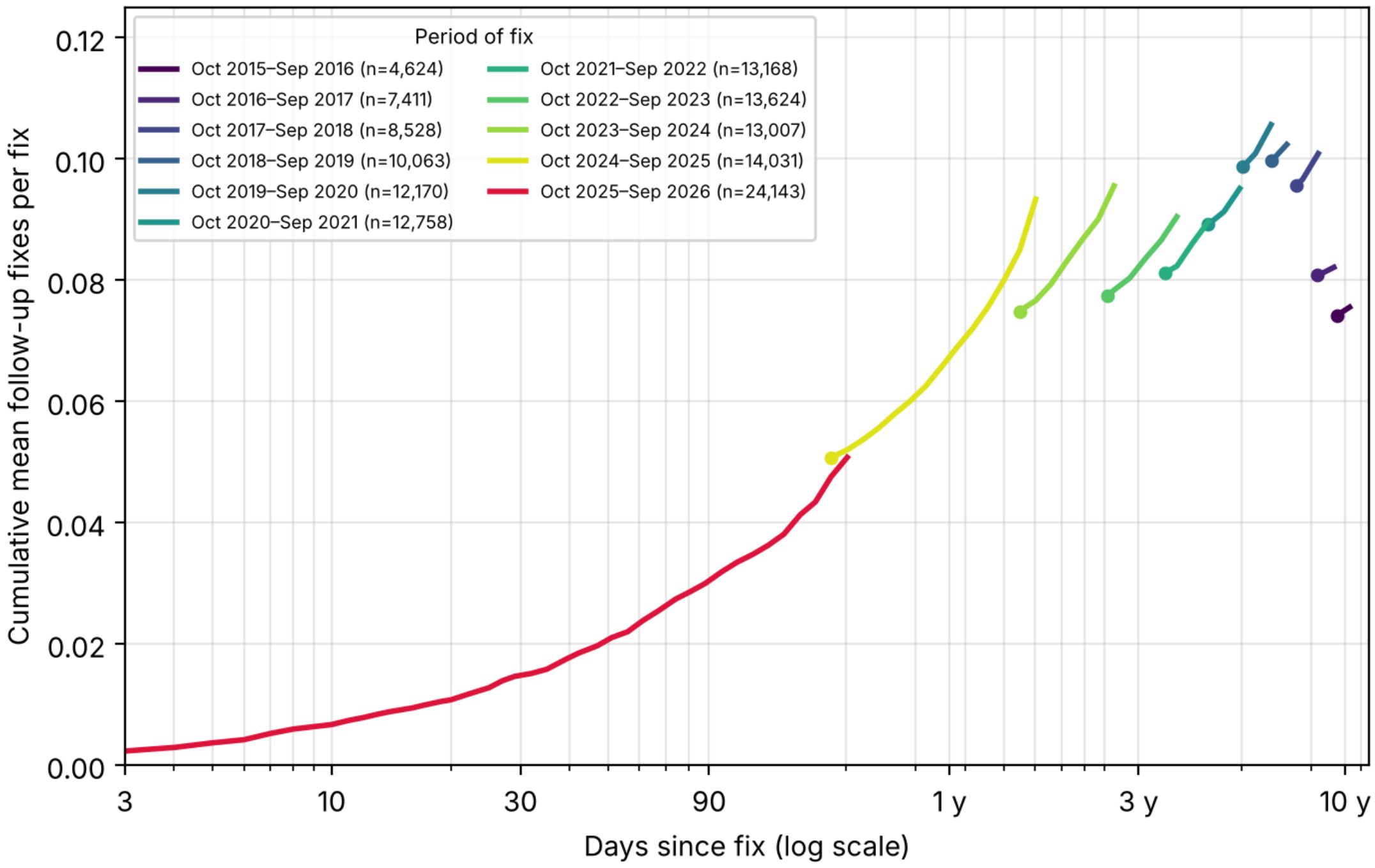

*Figure 7. The cohort curves of Figure 4, showing only the part from 1 October 2025.*

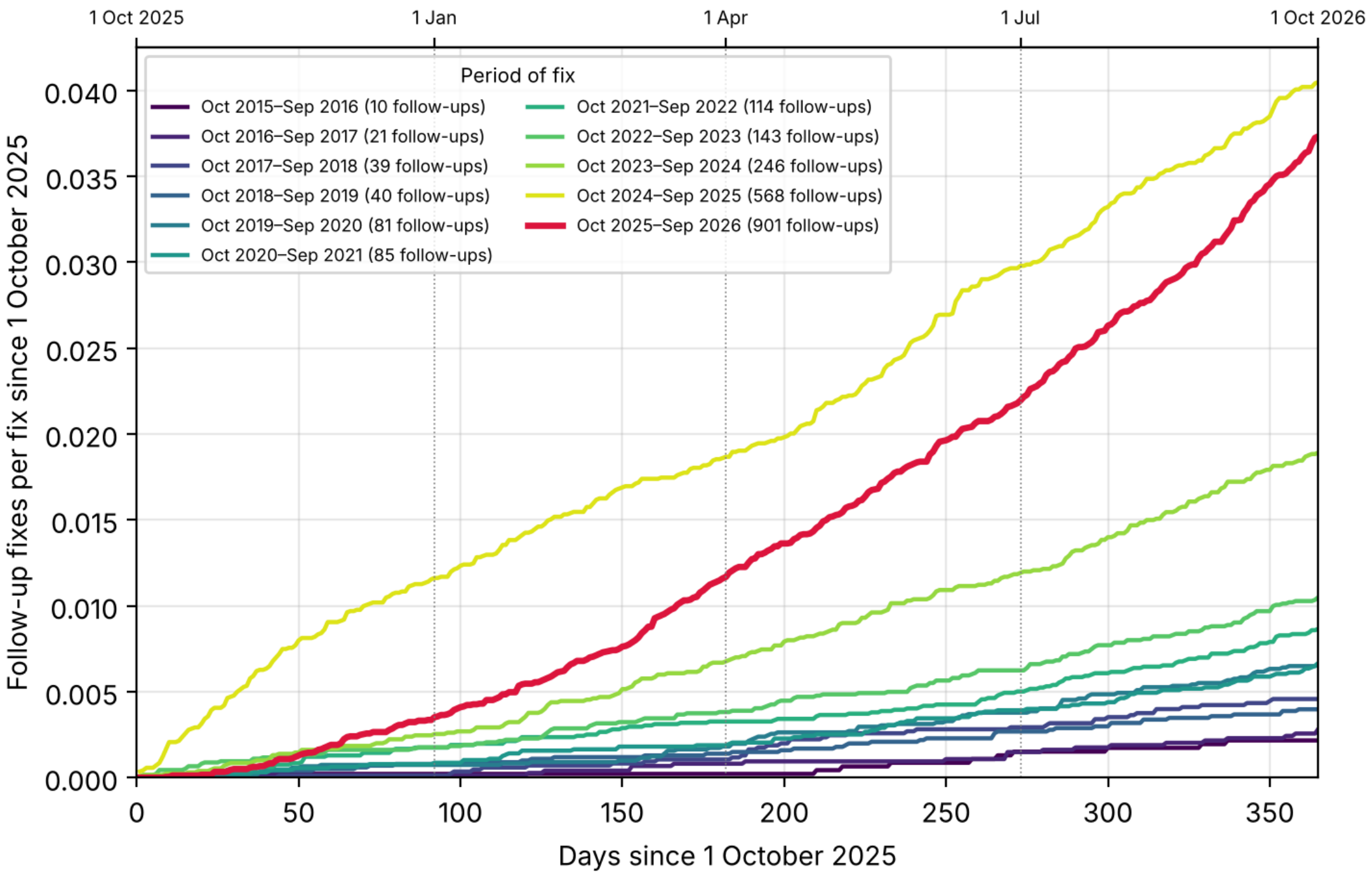

*Figure 8. Follow-ups per fix written since 1 October 2025, by year of fix, against days since that date. The red line is the cohort of fixes made during the window. Counts in the legend are for fixes of measurable size and are slightly below those of Table 6, which covers all tagged fixes.*

*Table 6. Multiplier in each half of October 2025–September 2026, by cohort of the commit fixed (intervals are 95% Poisson intervals conditional on the baseline, and so somewhat too narrow).*

| Commits made (Oct–Sep) | Ordinary: fixes received | Oct–Mar | Apr–Sep | Earlier fixes: follow-ups | Oct–Mar | Apr–Sep |
|---|---|---|---|---|---|---|
| 2015–16 | 643 | 1.5 ± 0.2 | 6.1 ± 0.5 | 10 | 0.3 ± 0.7 | 5.5 ± 3.6 |
| 2016–17 | 689 | 1.6 ± 0.2 | 4.4 ± 0.4 | 21 | 1.0 ± 0.8 | 2.6 ± 1.3 |
| 2017–18 | 704 | 1.7 ± 0.2 | 4.5 ± 0.4 | 39 | 1.3 ± 0.9 | 4.4 ± 1.6 |
| 2018–19 | 765 | 1.2 ± 0.2 | 4.3 ± 0.4 | 40 | 1.1 ± 0.6 | 2.7 ± 1.1 |
| 2019–20 | 1,119 | 1.9 ± 0.2 | 3.8 ± 0.3 | 81 | 1.3 ± 0.6 | 3.5 ± 0.9 |
| 2020–21 | 1,303 | 1.4 ± 0.2 | 4.5 ± 0.3 | 87 | 1.3 ± 0.5 | 3.3 ± 0.8 |
| 2021–22 | 1,326 | 1.3 ± 0.1 | 3.0 ± 0.2 | 117 | 1.4 ± 0.4 | 2.9 ± 0.7 |
| 2022–23 | 1,817 | 1.3 ± 0.1 | 2.9 ± 0.2 | 146 | 1.2 ± 0.3 | 2.4 ± 0.5 |
| 2023–24 | 2,435 | 1.4 ± 0.1 | 2.8 ± 0.1 | 257 | 1.4 ± 0.3 | 3.1 ± 0.5 |
| 2024–25 | 4,627 | 1.2 ± 0.1 | 2.5 ± 0.1 | 576 | 1.2 ± 0.1 | 3.0 ± 0.3 |
| 2025–26 (made in window) | 5,168 | 1.1 | 1.2 | 910 | 1.0 | 0.8 |

Every cohort shows the rise, and in the same months. The multiplier is not identical across cohorts. For ordinary commits it increases with the age of the code: 2.5–3.0 for code one to four years old and 4–6 for code five to ten years old. The joint model of Appendix C confirms that in 2026 the rate rose more for older code. The tagging correction was estimated for fixes in general and so supports the overall rise more firmly than it does this gradient. For earlier fixes the intervals are too wide to show a gradient. The separable description — a rate proportional to $s^{\gamma}$, falling as $1/a$ with age $a$, times a calendar-time multiplier $M(t)$ — holds well until 2026 and is a first approximation thereafter.

## 6.4 Who is producing the fixes

The kernel's guidance on AI coding assistants was written on 23 Dec 2025 and merged on 6 Jan 2026. It asked that AI-assisted contributions carry an `Assisted-by:` tag naming the tool and model; a procedure for AI bug-finding was added in August 2026; and the references to the tag were removed on 13 October 2026. The commit that removed them records that about 10% of commits in the 7.3 development cycle carried the tag "while actual LLM use is far more widespread", and that maintainers had concluded it gave no reliable signal. Declarations are therefore a lower bound, shaped by a policy that changed during the period, and a tag may refer to finding a bug, locating its origin, writing the patch or reviewing it. With that caution, Table 7 classifies the fixes to code more than 180 days old.

*Table 7. Fixes to code more than 180 days old, by author history and declared AI involvement.*

| Author | Oct–Dec 2025 | Jul–Sep 2026: declared AI | Jul–Sep 2026: undeclared | Jul–Sep 2026: all |
|---|---|---|---|---|
| Established | 1,500 | 783 | 2,105 | 2,888 |
| Occasional | 637 | 598 | 1,548 | 2,146 |
| First-time | 258 | 1,076 | 2,073 | 3,149 |
| All | 2,395 | 2,457 | 5,726 | 8,183 |

An established author has at least five tagged fixes in the two years before the window; an occasional author has fewer but some earlier commit in the kernel (a median of six, so the class includes a few experienced developers); a first-time author has none. Only six fixes declared AI involvement in the earlier quarter. Three points follow.

1. Fixes by first-time authors rose twelvefold, from 258 to 3,149, and are now 38% of the total. This is a change at the extensive margin: many more people are able to contribute fixes to old kernel code.
2. Established authors nearly doubled their output, and 27% of their fixes now declare AI. Their undeclared fixes rose by 40%.
3. Fixes declaring AI are 30% of the total and are numerically equivalent to 42% of the increase; excluding them entirely leaves a multiplier of about 2.7. They are no smaller than other fixes (median 7 source lines against 7–8), so the pattern is not an artefact of AI-assisted work being split into smaller patches.

The evidence that bears on discovery specifically is in the credits: of the thirty reporters most often credited in April–September 2026, twenty-five had no credited report in the previous year, and they include automated systems and security research groups.

## 6.5 Where the surge falls, and on what

The rise is not confined to peripheral code. Table 8 counts fixes to code more than 180 days old by the area of the kernel the fix touches.

*Table 8. Fixes to code more than 180 days old, by area: an average quarter of October 2024–September 2025 against July–September 2026.*

| Area | Per quarter before | Jul–Sep 2026 | Growth | With a CVE: before | Jul–Sep 2026 |
|---|---|---|---|---|---|
| Filesystems | 178 | 901 | 5.1 | 32% | 47% |
| Core kernel | 79 | 380 | 4.8 | 30% | 52% |
| Networking core | 187 | 857 | 4.6 | 49% | 75% |
| Memory management | 58 | 220 | 3.8 | 27% | 22% |
| All core areas | 553 | 2,514 | 4.5 | 36% | 54% |
| Drivers | 1,211 | 4,116 | 3.4 | 22% | 36% |
| Architecture code | 268 | 713 | 2.7 | 9% | 25% |
| All | 2,281 | 8,183 | 3.6 | 22% | 38% |

“All core areas” adds the block layer, security, crypto, library and io_uring code to the four areas listed. Tables 7 and 8 count fixes, each once, including the 17% whose target predates October 2015; Table 5 counts fix-to-target links for code written from October 2015. Core areas, which are the most heavily fuzzed and reviewed, grew faster than drivers. Nor is the surge visibly made up of low-value fixes. The share of these fixes that received a CVE rose from 22% to 38%, about 500 a quarter to about 3,100. Fixes whose message mentions a use-after-free rose from 78 a quarter to 607, and an out-of-bounds access or overflow from 96 to 712, in each case doubling as a share of the total, while the share mentioning a null dereference fell. Declared AI assistance is evenly spread, at 28–38% in the main core areas and 29% in drivers. These are proxies: the kernel assigns CVEs to any fix with a plausible security consequence, and none of them measures exploitability. They are also growth rates in completed fixes, not adjusted for the size and age of each area’s code, so differences between areas may partly reflect where researchers chose to look.

## 6.6 Growth, and what is being discovered

Fitted to weekly counts for 2026, three forms for $M(t)$ are statistically indistinguishable: exponential growth of 0.3–0.4% a day (doubling in 170–220 days), linear growth of about 1.0 per 100 days, and a logistic curve levelling near 5. A constant is rejected ($\Delta\mathrm{AIC} > 80$). By the end of 2027 the three give multipliers of about 25, 8 and 5 respectively. Further months of data should discriminate between them, unless discovery or remediation capacity itself changes.

The contents of a commit made years ago are fixed, so a rise in the rate at which it is found defective cannot reflect a fall in its quality; like a more sensitive diagnostic test, it reflects better detection. (Later changes to surrounding code can expose a previously harmless defect, which is the one qualification.) If a commit of size $s$ contains on average $B(s)$ latent defects, each with detection intensity $h$, its expected discoveries by age $a$ are

$$D(s,a) = B(s)\left[1 - \exp\left(-\int_0^a h(u)\,du\right)\right]. \qquad [2]$$

Historical rates identify neither $B$ nor $h$ separately. The $1/a$ law would arise if defects had detection intensities spread evenly on a logarithmic scale over several orders of magnitude. The 2026 shock then carries information about $B$. That code eight to ten years old responds at least as strongly as recent code shows that the historical process had not exhausted the discoverable stock in the oldest cohorts. The stock may be both finite and much larger than the old curves implied, and what follows depends on three things that can operate together.

1. *Depletion*. The rate peaks and declines although discovery capability is sustained.
2. *Continuing improvement in discovery*. The rate goes on rising, possibly masking depletion.
3. *Saturation of capacity*. The rate levels off although a substantial stock remains, because discovery or remediation effort is limited.

No cohort yet shows a peak. That the multiplier is larger for older code suggests that the new methods are disproportionately effective on defects with the lowest historical detection intensities. One simple form of that hypothesis is that an intensity $h$ becomes $A\,h^{\beta}$ with $\beta < 1$, so that the multiplier $A\,h^{\beta-1}$ is greatest where $h$ was smallest; old cohorts, which by now consist mostly of hard-to-find defects, would then respond most. The evolution of the cohort multipliers over 2027 will test it, and provide important insights on the likely trends.

## 6.7 New fixes, and review

Against this, fixes made during the window seem to have attracted fewer follow-ups than normal: 1.0 times normal in the first half and 0.8 in the second, where new ordinary commits ran at 1.1 and 1.2. Follow-ups within 90 days per fix fell from 0.044 (2025) to 0.025 (May–August 2026) for CVE fixes and from 0.039 to 0.022 for other tagged fixes. The fall is not an artefact of follow-ups still awaiting a merge window at the snapshot. Follow-ups within 30 days per fix were 0.021–0.024 for the fixes of March, June and September 2025 and 0.010, 0.014 and 0.012 for those of May, June and July 2026, which had four to six months to reach mainline; such follow-ups are committed a median of one to five days after they are written. In absolute terms the number of these quick follow-ups did not change: about 29 a month in 2025 and 24–37 a month from May to September 2026, while fixes doubled and tripled. Either quick detection has a roughly fixed capacity, or the additional fixes are of a kind that seldom needs a quick follow-up. Three explanations are open.

1. *Deferred detection*. Attention has moved to old code: fixes to targets under 180 days old stayed at 400–600 a month while fixes to targets over five years old rose from about 250 to about 1,400.
2. *Better or easier fixes*. The fixes of 2026 are disproportionately for old defects, which may be more isolated, and many were written or checked with AI assistance.
3. *Composition or recording*. A change in who writes fixes or in how reliably follow-ups carry the tag.

These predict different things at longer horizons: under the first, follow-ups to the fixes of mid-2026 recover to the historical level within one to two years; under the second they remain lower. The 180-day windows for June and July 2026 close in late December 2026 and late January 2027, and the 365-day windows in mid-2027.

Table 9 gives what the commit record shows about review of fixes to old code.

*Table 9. Recorded review of fixes to code more than 180 days old, six-month periods.*

| | Apr–Sep 2025 | Apr–Sep 2026 | Change |
|---|---|---|---|
| Fixes | 4,498 | 13,562 | +202% |
| Distinct authors | 1,522 | 2,682 | +76% |
| Fixes with a `Reviewed-by:` tag | 2,451 (54%) | 6,601 (49%) | +169% |
| Distinct reviewers | 785 | 1,188 | +51% |
| Review tags per reviewer | 4.2 | 7.2 | +71% |
| Fixes with a `Tested-by:` tag | 530 (11.8%) | 932 (6.9%) | +76% |

Recorded review nearly kept pace with a tripling of fixes: half as many reviewers again, each reviewing 70% more. Testing did not: tested fixes rose by three-quarters and coverage fell from 12% to 7%. Merge latency did not lengthen (median 4.8 days from authorship to commit in 2025, 5.2 in May–October 2026). A maintainer may review without adding a tag, so these are measures of recorded review. They describe falling coverage under rising throughput, consistent with pressure on capacity and not yet with its exhaustion.

# 7 A model of remediation

In this section we provide a simple model of remediation and explore how far we are away from a situation in which the behaviour exhibits undesirable properties such as fixes becoming self-perpetuating. The model is stationary and is used here for scenarios. With a multiplier $M(t)$ that changes over time the re-fix ratio of a fix depends on the date it was made.

$$R_H(t) = \int_0^H M\,(t + a)\, g(a)\, da \qquad [3]$$

with $g$ the historical intensity, and a full treatment would use an age-dependent branching process with a calendar-dependent kernel.

## 7.1 Cascades

Let a discovered bug require a fix of size $S$ with mean $\mu$. Let the fix generate $N$ follow-ups with conditional mean $n(s) = E[N \mid S = s]$ and variance $v(s)$, each follow-up an independent fix from the same size distribution with its own follow-ups. The re-fix ratio is $R = E[n(S)]$, and for $R < 1$ the expected total work per discovered bug, over all generations, is

$$w = E[W] = \frac{\mu}{1-R}. \qquad [4]$$

From $W = S + \sum_{i=1}^{N} W_i$ and the law of total variance,

$$\mathrm{Var}(W) = \frac{w^2\, E[v(S)] + \mathrm{Var}(S + w\, n(S))}{1-R}. \qquad [5]$$

This holds for any offspring law. For the empirical $n(s) = c_0 + c_1 s^\gamma$ with negative-binomial variance $v = n + \alpha n^2$, all the moments can be computed from the observed sizes. Fitted to the tagged fixes of 2018–2023 at the two-year horizon, $c_0 = 0.007$, $c_1 = 0.012$, $\gamma = 0.72$ and $\alpha = 2.6$; with $\mu = 14.4$ and $\mathrm{CV}_S^2 = 4.1$ the result is $R = 0.076$, $w = 15.6$ lines and $\mathrm{CV}_W^2 = 4.3$. Cascades amplify variability as well as the mean, without bound as $R \to 1$: if all follow-up rates were scaled by 2.5, 5 and 8, $R$ would be 0.19, 0.38 and 0.61 and $\mathrm{CV}_W^2$ 4.7, 6.4 and 12.

The horizon applies to each generation: $R_H$ counts the follow-ups a fix attracts within $H$ of its own date, so $w$ is the total work of a cascade in which every fix has an offspring window of length $H$, not the work done within $H$ of the original bug. How quickly that work arrives after a shock would need the renewal equation

$$W(t) = \mu + \int_0^t k(a)\, W(t-a)\, da \qquad [6]$$

with $k$ the offspring intensity at age $a$.

A branching process with $R \leq 1$ dies out with certainty, though at $R = 1$ its expected size is infinite; only for $R > 1$ is there a positive probability that a cascade never ends.

## 7.2 How far is the kernel from critical?

On pre-2026 rates a fix attracts about 0.05 follow-ups in its first 180 days and a further 0.04 by five years, so $R_{5y} \approx 0.09$. The 2026 multiplier applies to the second part. Table 10 gives the result if a multiplier $M$ were sustained and discovery did not deplete the stock of defects — the second reading of Section 6.6. The lifetime ratio is not identified.

*Table 10. Follow-ups per fix under a sustained multiplier on code more than 180 days old, without depletion.*

| Multiplier $M$ | $R$ at 5 years | $R$ at 10 years |
|---|---|---|
| 1 (2018–2025) | 0.09 | 0.10 |
| 3 (third quarter of 2026, corrected) | 0.17 | 0.20 |
| 7 | 0.33 | 0.39 |
| 14 | 0.61 | 0.73 |

At the present level the five-year ratio nearly doubles and expected work per bug rises by 10%. The ratio approaches 1 only for a sustained multiplier of 20–25.

Size works in the same direction. Because the re-fix law is concave, $R$ depends on the whole size distribution and not on its mean alone: by Jensen's inequality $R \leq c_0 + c_1\mu^\gamma$, so the process is subcritical whenever

$$\mu < \left((1-c_0)/c_1\right)^{1/\gamma} \qquad [7]$$

about 470 lines on the coefficients of Section 7.1. If sizes are scaled by a common factor with the kernel's shape preserved, $R$ reaches 1 at a mean of about 600 lines. Both figures use pre-2026 coefficients, and extrapolate far beyond the kernel's mean of 14 and beyond the range in which the law is measured (Section 5.1); with a multiplier above 1 they would be lower, and lower still if the size exponent has risen (Section 6.3). A consequence of concavity is that a mean-preserving spread of fix sizes generates fewer follow-ups while imposing a larger queueing penalty.

## 7.3 Demand and queues

Suppose bugs arrive at rate $\lambda$ and remediation capacity is $c$ lines per unit time. Treating each cascade as one job, utilisation is

$$\rho = \frac{\lambda\,\mu}{c\,(1-R)}, \qquad [8]$$

and for Poisson arrivals and a single server the mean wait is

$$E[T_q] = \frac{\rho}{1-\rho}\cdot\frac{w}{c}\cdot\frac{1+\mathrm{CV}_W^2}{2}. \qquad [9]$$

The multiplier acts twice: on $R$, and on the arrival of primary discoveries, but only on that part of them which concerns code more than six months old. Before 2026 a share $p \approx 0.63$ of fixes to ordinary code had targets of that age (0.60 over 2018–2025). With $R(M) = 0.05 + 0.04M$ at five years, demand relative to the historical level is

$$\frac{L(M)}{L(1)} = \frac{0.91\,[1+p\,(M-1)]}{0.95-0.04\,M}. \qquad [10]$$

*Table 11. Remediation demand relative to its historical level, no-depletion scenario, $p = 0.63$.*

| Multiplier $M$ | 1 | 2 | 3 | 5 | 7 | 10 | 14 |
|---|---|---|---|---|---|---|---|
| Re-fix ratio at 5 years | 0.09 | 0.13 | 0.17 | 0.25 | 0.33 | 0.45 | 0.61 |
| Demand | 1.0 | 1.7 | 2.5 | 4.3 | 6.5 | 11 | 21 |

Demand at the present multiplier is about two and a half times the historical level, and the practical difficulty arises long before the branching process approaches criticality: a re-fix ratio of 0.4 is comfortably subcritical and corresponds to a ninefold demand. Here $\lambda$ counts primary discoveries only; follow-ups enter through $R$. Waiting time is more sensitive still. As an illustration only — the kernel's utilisation is not measured — a system at $\rho = 0.3$ whose demand rises two and a half times moves to $\rho = 0.75$, and its mean wait rises sevenfold with the distribution of service times held fixed, and by more if cascades lengthen. The variability factor $(1 + \mathrm{CV}_W^2)/2$ is 2.6 for kernel commits and about 8 if the unit of work is the patch series (Section 4).

Two features of this abstraction should be kept in view. First, follow-ups become discoverable months or years after the fix, so a cascade re-enters the queue with delay, as in a Hawkes process; treating it as one job exchanges that clustering of arrivals for extra variance in service time. Second, capacity is expressed in lines, which is a proxy: if a five-line concurrency fix takes as long to review as fifty lines of routine change, the variability of remediation time is not that of commit size. Nor does a single-class queue predict which follow-ups are deferred under load; the deferred-detection hypothesis of Section 6.7 amounts to a detection function for recent code that shifts later as utilisation rises.

## 7.4 Granularity

If the empirical law is unchanged by subdivision, splitting a remediation of size $S$ into $k$ equal patches changes expected direct follow-ups from $c_0 + c_1 S^\gamma$ to $k\,c_0 + k^{1-\gamma} c_1 S^\gamma$: with $\gamma = 0.7$, ten patches attract about twice the follow-ups of one. Against that, smaller units reduce queueing variability. The law was estimated on naturally occurring commits, and deliberate subdivision may change how a change is reviewed or how its defects are attributed, so this is a prediction to be tested, not a result. Any method that bounds the scope of a change — modular design, or compositional verification where a defect can be corrected without changing a module's contract — acts on the same trade-off.

# 8 Discussion

**What the kernel showed until 2026.** Over a decade, in the most scrutinised codebase there is, fixes stayed small, each attracted about 0.08 further fixes within two years, the need grew less than proportionally with size, and none of this was improving. Discovery difficulty and remediation work are unrelated: a ten-year-old bug costs the same seven lines.

**What 2026 shows.** Those rates measured the pace of discovery. When discovery accelerated, code more than six months old yielded defects at about three times its historical rate within nine months, older code more than newer, with no sign yet of exhaustion, and fixes yielded them at the same multiple as the code around them. A fix is not the end of a defect's history; it is new code with the same exposure as any other. The quantity that matters for the sustainability of patching is the stock of latent defects per line, which the historical record does not reveal and which better discovery keeps revising upward.

**What has changed, and what part is AI.** The volume of fixes, by a factor of two to three; the age of the bugs fixed; and the rate at which old code is found wanting, by a factor of about three and rising. Not the size of fixes. The change dates from the weeks in which the kernel adopted guidance for AI coding assistants; three-tenths of the fixes concerned declare AI assistance; and most of the rest of the increase comes from authors and reporters new to the kernel. That is strong circumstantial evidence, not proof, that AI-assisted discovery is the cause. Whether AI has improved the quality of fixes is the open question of Section 6.7. The claim that defence will keep pace because AI also writes fixes is a claim about the re-fix ratio and about capacity: this study finds the first rising for all existing code, and recorded testing falling well behind throughput.

**Limitations and further work.**

1. *Arrivals*. The study observes completed fixes. Arrival dates for publicly reported bugs are in the kernel mailing-list archive; those for privately reported vulnerabilities are held by the kernel security team and by national cyber-security authorities. With either, the queue itself could be measured rather than inferred.
2. *Patch series*. The series proxy should be replaced by series identifiers from the mailing-list archive.
3. *Mechanism*. A stratified manual classification of a few hundred follow-ups, as incomplete, regressive or other, would test whether size-dependent and size-independent components have different causes.
4. *Other codebases*. Provisional measurements on TensorFlow, Django and OpenClaw give median security fixes of 10, 34 and 66 lines, but language, maturity, conventions and selection are confounded, and no re-fix law has been estimated for them.
5. *Bug age as difficulty*. An old bug may simply lie on a rarely executed path; code age and test coverage should be controlled for.
6. *Tag coverage*. Follow-ups without a `Fixes:` tag are missed, and coverage may vary under load. The share of commits carrying the tag rose to 31% in 2026, which reflects the surge in fixing; the probability that a fix carries the tag rose by a factor of 1.12–1.15 (Appendix C).
7. *AI attribution*. Declaration of AI assistance followed a policy that was introduced and withdrawn within the period, does not say what the assistance was, and is matched here by keyword with a false-positive rate of about 0.3%.
8. *The baseline*. "Normal" rates are estimated from 2018–2025 and assumed to depend on age alone; age, period and cohort effects cannot all be identified without that assumption, which the seven stable years support.
9. *Attrition*. Exposure is counted per commit with no adjustment for code since rewritten or deleted. Less old code survives to be found defective, which probably makes the multiplier for old code conservative, though that depends on how survival differed between the baseline and recent periods.
10. *Severity*. Fixes are counted equally; CVE assignment and keywords are the only indicators of security value used.
11. *Lines as work*. The finding that fix size is independent of bug age concerns lines changed, not developer time.

# Appendix A. The distribution of fix size

Fitted by discrete maximum likelihood on the 18,358 kernel CVE fixes that remain when all deletion-dominated commits are excluded; under the rule used in the text (18,704 fixes) the log-logistic parameters are unchanged to two figures ($a = 7.1$, $b = 1.49$).

*Table A1. Candidate laws for* $\ln S$.

| Family of $\ln S$ | Parameters | $\Delta \log L$ vs log-normal | Predicted fixes >300 lines (25 observed) |
|---|---|---|---|
| Normal (log-normal) | 2 | 0 | 13 |
| Logistic (log-logistic) | 2 | −294 | 70 |
| Student-t (log-t) | 3 | 0 | 13 |
| Asymmetric Laplace (double Pareto) | 3 | −938 | 377 |
| Gumbel (Fréchet in $s$) | 2 | +85 | hundreds |

The log-t collapses onto the log-normal, since tails polynomial in $\ln s$ are heavier than any power of $s$ and the data reject them. The log-normal beats the log-logistic at 2 lines and at 20–100 lines and loses at 3–10 and above 100. Within each class of bug (races, leaks, null dereferences and so on) the picture is the same, so the shape is not a mixture effect.

*Table A2. Truncated-Pareto exponent on* $[L, U]$ *lines.*

| Range | n | Exponent (95% CI) |
|---|---|---|
| 20–300 | 3,843 | 1.30 (1.26–1.35) |
| 30–300 | 2,382 | 1.50 (1.42–1.57) |
| 40–300 | 1,579 | 1.59 (1.49–1.69) |
| 30–1,000 | 2,406 | 1.60 (1.54–1.66) |
| 50–1,000 | 1,146 | 1.82 (1.71–1.93) |

The exponent rises with the lower threshold throughout, which is the signature of a curved (log-normal-like) tail rather than a power law. An earlier draft's "exponent 1.5" was an average over this curvature.

# Appendix B. Sensitivity to the exclusion of removals

| Rule | n | Median | Mean | Squared coefficient of variation |
|---|---|---|---|---|
| No exclusion | 18,725 | 7 | 27 | 2,100 |
| Exclude deletion-dominated commits above 100 lines (used) | 18,704 | 7 | 15.7 | 3.9 |
| Exclude all deletion-dominated commits | 18,358 | 7 | 15.8 | 3.9 |

Of the 367 deletion-dominated CVE fixes, 303 are 20 lines or fewer and are ordinary repairs; the 21 above 100 lines are removals of components and account for the whole difference in the mean and second moment.

# Appendix C. Robustness of the 2026 multiplier and of the re-fix law

**C1. Tagging.** Two sets of fixes can be identified without the `Fixes:` tag: commits marked for stable backporting, and commits whose subject says "fix" and does not concern typos, comments, documentation, warnings or build errors.

| Period | Stable-marked: per month | Share tagged | Subject says fix: per month | Share tagged |
|---|---|---|---|---|
| Oct 2018–Sep 2021 | 315 | 64% | 824 | 52% |
| Oct 2021–Sep 2024 | 339 | 71% | 765 | 63% |
| Oct 2024–Sep 2025 | 420 | 81% | 705 | 69% |
| Jan–Mar 2026 | 572 | 84% | 947 | 73% |
| Apr–Jun 2026 | 987 | 89% | 1,190 | 76% |
| Jul–Sep 2026 | 1,441 | 93% | 1,380 | 77% |

Between the year to September 2025 and the third quarter of 2026 the tagged share rose by a factor of 1.15 on the first measure and 1.12 on the second. The correction is applied to the multiplier measured against that same year (3.42), giving about 3.0. The share had also been rising within the 2018–2025 baseline, which is why the like-for-like comparison uses the latest year only. This test measures tagging for fixes in general; whether fixes to very old code were historically tagged less often than others cannot be tested without the tag itself.

**C2. Variants of the multiplier.** All code written Oct 2015–Sept 2025, ages > 180 days, Q3 2026.

| Variant | Multiplier |
|---|---|
| A. Baseline October 2018–September 2025 | 3.89 (3.32–4.54) |
| B. Baseline October 2024–September 2025 | 3.42 (2.86–3.84) |
| C. B with tagging correction (÷1.12–1.15) | 3.0 (2.5–3.4) |
| D. A on commit dates | about 3.8 |
| E. A excluding fixes declaring AI | about 2.7 |
| F. C excluding fixes declaring AI | about 2.1 |

Intervals are from 400 bootstrap replications that resample the weeks of the quarter and the years of the baseline (for B, Poisson resampling of baseline counts); they are two to three times wider than Poisson intervals conditional on the baseline. Variants D to F were computed on a pooled baseline and are scaled to A.

**C3. Baseline period and age bands.** Normal rates are estimated in fourteen age bands from 0 to 20 years, using commits back to 2006 so that every age reached in the window is covered. For earlier fixes the third-quarter multiplier is 3.53, 3.55 and 3.51 on baselines of Oct 2018–Sept 2025, 2018–2024 and 2021–2025. For ordinary commits it is 3.92, 4.01 and 3.58; the yearly multiplier for ordinary commits rose from 0.83 in 2018–19 to 1.14 in 2024–25, so the most recent baseline is the conservative one. An earlier version of this analysis used finer age bands and only commits since October 2015; it left the highest ages without baseline exposure and gave unstable figures for the oldest cohorts.

**C4. Joint model.** A Poisson regression of follow-up counts on log size, age band, calendar period (years 2019–2025 and the first three quarters of 2026) and population (tagged fixes; the sample of ordinary commits), with exposure as offset, over 759 cells and 17,964 events.

| Quantity | Estimate |
|---|---|
| Size exponent $\gamma$ | 0.712 (s.e. 0.006; about 0.01 allowing for over-dispersion) |
| Fixes relative to ordinary commits, at equal size and age | 1.14 |
| Period effects 2020–2025 relative to 2019 | 0.83–0.98 |
| Old code (over 180 days), 2026 Q1, Q2, Q3 relative to 2019–2025 | 1.37, 1.98, 3.13 |
| Code under 180 days relative to old code, 2026 Q1, Q2, Q3 | 0.82, 0.44, 0.23 |
| 2026 × log size | +0.16 (s.e. 0.014) |

| Quantity | Estimate |
|---|---|
| 2026 × age band, relative to 180–365 days | 0.37 (0–30 d), 0.67 (30–180 d), 1.48 (1–3 y), 1.44 (3–5 y), 1.71 (over 5 y) |

Residual deviance is about 2.2 times its degrees of freedom both before 2026 and overall, so standard errors should be inflated by about 1.5. The separable model fits the years before 2026 as well as it fits the whole; the 2026 effect is not purely multiplicative, being larger for older commits and, possibly, though this is less clear, for larger ones (C8). The third-quarter figure here (3.1) is lower than in Table 5 because the model's sample of ordinary commits is younger (2021–2025).

**C5. Count model.** For CVE fixes at 180 days the linear-mean Poisson model has residual variance-to-mean ratio 1.40; a negative binomial with variance $\mu + \alpha\mu^2$ gives $\alpha = 3.3$ and an AIC lower by 29. For tagged fixes of 2018–2023 at two years, with the power-law mean, $\alpha = 2.6$. Confidence intervals for the coefficients from a bootstrap clustered on 363 subsystems are within 5% of the unclustered ones. A profile-likelihood interval for $\gamma$ on CVE fixes alone is 0.70–0.93.

**C6. Offspring by size class.** Mean follow-ups within 180 days to a CVE fix, by size of parent (rows) and of follow-up (columns).

| Parent | Small (≤10) | Medium (11–50) | Large (>50) |
|---|---|---|---|
| Small | 0.024 | 0.008 | 0.001 |
| Medium | 0.053 | 0.026 | 0.006 |
| Large | 0.133 | 0.068 | 0.037 |

The Spectral radius is 0.066 from unrounded entries versus a single-type ratio 0.061, so the dependence of follow-up size on parent size adds little to the re-fix ratio, and provides little evidence that the size of the 'offspring' depends on the size of the parent.

**C7. Cohort heterogeneity.** Separate exponential growth rates for the 2019–2025 cohorts of fixes are not rejected in favour of a common rate ($\chi^2 = 8.6$ on 6 degrees of freedom, $p = 0.20$), which is a failure to reject, not a demonstration of equality; Table 6 shows that for ordinary commits the multiplier differs between older and newer cohorts.

**C8. Size and the 2026 rise, with controls.** Targets are commits of known size written before April 2024: tagged fixes from Oct 2015 and the random sample of ordinary commits from 2021 (124,106 commits). Each had the same exposure, at ages over 180 days, in the baseline year (Oct 2024–Sept 2025) and in the third quarter of 2026. For the 1,662 fixes they received in those periods (1,084 and 578), a logistic regression of period on log size estimates the change in the size exponent directly.

| **Model** | **Change in exponent (s.e.)** |
|---|---|
| Size only | +0.06 (0.03) |
| + population | +0.08 (0.04) |
| + year written, subsystem, author experience | +0.09 (0.04) |
| Earlier fixes only | +0.09 (0.06) |
| Ordinary commits only | +0.07 (0.05) |
| Core subsystems | +0.24 (0.08) |
| Drivers | +0.01 (0.06) |
| Fixes by established authors declaring no AI | +0.04 (0.07) |
| Fixes declaring AI (against all baseline fixes) | +0.11 (0.06) |

The sample is mostly code of 2021–24, so its raw multiplier (2.1) is below that of Table 5.